\documentclass[runningheads]{llncs}
\usepackage[T1]{fontenc}
\usepackage{cite}
\usepackage{amsmath,amssymb,amsfonts}
\usepackage{algorithmic}
\usepackage{graphicx}
\usepackage{textcomp}
\usepackage{xcolor}
\usepackage{rotating}
\usepackage{cite}
\usepackage{amsmath,amssymb,amsfonts}
\usepackage{algorithmic}
\usepackage{graphicx}
\usepackage{textcomp}
\usepackage{xcolor}
\usepackage{cite}
\usepackage{amsmath,amssymb,amsfonts}
\usepackage{algorithmic}
\usepackage{textcomp}
\usepackage{xcolor}
\usepackage{booktabs}
\usepackage{multirow}
\newcommand{\etal}{\emph{et al.}}
\usepackage{algorithm}
\usepackage{algorithmic}
\usepackage{multirow}
\usepackage{dsfont}
\usepackage{graphicx}
\usepackage{subcaption}
\usepackage{hyperref}

\usepackage{graphicx}
\usepackage{color}

\begin{document}
\title{LLM-Driven Medical Document Analysis: Enhancing Trustworthy Pathology and Differential Diagnosis}

\titlerunning{LLM-Driven Medical Document Analysis}

\author{Lei Kang$^\dag$\orcidID{0000-0002-1962-3916} \and\\
Xuanshuo Fu$^\dag$\orcidID{0000-0002-0377-2263} \and\\
Oriol Ramos Terrades\orcidID{0000-0002-3333-8812} \and\\
Javier Vazquez-Corral\orcidID{0000-0003-0414-7096} \and\\
Ernest Valveny\orcidID{0000-0002-0368-9697} \and\\
Dimosthenis Karatzas\orcidID{0000-0001-8762-4454}
}
\authorrunning{L. Kang et al.}
%
\institute{Computer Vision Center, Universitat Autònoma de Barcelona, Barcelona, Spain\\
\email{\{lkang, xuanshuo, oriolrt, javier.vazquez, ernest, dimos\}@cvc.uab.es}}

\maketitle              

{\renewcommand\thefootnote{}\footnotetext{$^\dag$ These authors contributed equally to this work.}}

\begin{abstract}

Medical document analysis plays a crucial role in extracting essential clinical insights from unstructured healthcare records, supporting critical tasks such as differential diagnosis. Determining the most probable condition among overlapping symptoms requires precise evaluation and deep medical expertise. While recent advancements in large language models (LLMs) have significantly enhanced performance in medical document analysis, privacy concerns related to sensitive patient data limit the use of online LLMs services in clinical settings. To address these challenges, we propose a trustworthy medical document analysis platform that fine-tunes a LLaMA-v3 using low-rank adaptation, specifically optimized for differential diagnosis tasks. Our approach utilizes DDXPlus, the largest benchmark dataset for differential diagnosis, and demonstrates superior performance in pathology prediction and variable-length differential diagnosis compared to existing methods. The developed web-based platform allows users to submit their own unstructured medical documents and receive accurate, explainable diagnostic results. By incorporating advanced explainability techniques, the system ensures transparent and reliable predictions, fostering user trust and confidence. Extensive evaluations confirm that the proposed method surpasses current state-of-the-art models in predictive accuracy while offering practical utility in clinical settings. This work addresses the urgent need for reliable, explainable, and privacy-preserving artificial intelligence solutions, representing a significant advancement in intelligent medical document analysis for real-world healthcare applications. The code can be found at \href{https://github.com/leitro/Differential-Diagnosis-LoRA}{https://github.com/leitro/Differential-Diagnosis-LoRA}.

\end{abstract}

\section{Introduction}

Medical document analysis is essential for extracting critical information from unstructured healthcare records, directly supporting tasks like differential diagnosis. Differential diagnosis aims to identify the most probable condition among multiple potential pathologies that often coexist in a single patient and exhibit overlapping symptoms. This complexity makes accurate diagnosis challenging, requiring physicians to meticulously distinguish between conditions using deep medical knowledge, critical thinking, and sometimes additional diagnostic tests. As the demand for precise and timely diagnoses grows, healthcare providers are increasingly leveraging advanced technologies like Large Language Models (LLMs) to enhance decision-making and patient outcomes. State-of-the-art models such as OpenAI's GPT-4~\cite{achiam2023gpt} have demonstrated impressive capabilities in text generation, language translation, and question answering, although detailed information about their training methodologies remains largely undisclosed. Open-source alternatives like Meta's LLaMA models~\cite{touvron2023llama}, ranging from 7B to 70B parameters, offer powerful options. Recently, Meta released LLaMA v3~\cite{dubey2024llama}, which scales up to 405B parameters, expanding applications across fields including healthcare. However, deploying LLMs in sensitive domains like healthcare raises significant concerns regarding privacy, transparency, and trust, especially when handling sensitive patient data. Privacy breaches in these contexts could lead to severe ethical and legal consequences. Therefore, while LLMs hold great potential for medical diagnosis, record summarization, and medical literature interpretation, the inherent risks associated with cloud-based deployment models highlight the need for locally deployable, trustworthy AI solutions that deliver high performance while ensuring strict privacy protections.

Two key tasks emerge as critical for healthcare AI systems: pathology prediction and differential diagnosis. Pathology prediction focuses on identifying the most likely disease based on patient data, while differential diagnosis involves distinguishing between various diseases or conditions that could present with similar symptoms. Both tasks require systems capable of providing multiple highly probable diagnostic outcomes based on a patient's symptoms and medical history, ensuring comprehensive and accurate pre-diagnosis.

In this work, we introduce an innovative medical document analysis method designed for trustworthy pathology prediction and differential diagnosis. Our approach utilizes a LoRA-tuned LLaMA-v3 LLM, specifically optimized to enhance diagnostic accuracy and reliability.  Our system processes patient data locally, eliminating the reliance on cloud-based LLM APIs and ensuring full control over sensitive medical information. Low-Rank Adaptation (LoRA)~\cite{hu2021lora} has demonstrated its effectiveness in fine-tuning large language models, enabling high task-specific performance while meeting stringent computational and privacy requirements. Utilizing the LoRA approach, we propose updating the self-attention modules in each layer of the LLaMA-v3 model. Using DDXPlus, the largest publicly available benchmark dataset for differential diagnosis, we demonstrate that our fine-tuned model surpasses existing methods in varied performance metrics. Additionally, we incorporate explainability techniques into the platform to ensure that predictions are robust, trustworthy, and transparent, which are key factors in gaining acceptance in clinical environments. Finally, we have developed a web-based interface that provides users with easy access to early-stage pre-diagnosis, helping to alleviate congestion in healthcare facilities.

This paper makes the following key contributions:

\begin{itemize}
    \item We introduce a novel LLM-driven approach to medical document analysis focused on pathology prediction and differential diagnosis. This method is incorporated into a new web-based medical pre-diagnosis platform that operates locally, making it well-suited for sensitive hospital settings.
    \item Extensive evaluations on the DDXPlus dataset show that our model outperforms state-of-the-art approaches, demonstrating strong potential for practical application in real-world medical diagnosis scenarios.
    \item We implement explainability techniques to enhance the platform's trustworthiness, offering clear insights into the model’s decision-making process to ensure transparency and reliability in clinical use.
\end{itemize}

\section{Related Work}

\textbf{LLM in Medical Document Analysis.}

The application of LLMs in healthcare has progressed rapidly, providing significant benefits in tasks like medical diagnosis~\cite{garg2023exploring,wang2024beyond,ullah2024challenges}, natural language processing of patient records~\cite{hossain2023natural,yuan2023llm,schmiedmayer2024llm}, and decision support in clinical settings~\cite{ferdush2024chatgpt,prabhod2023integrating,mcpeak2024llm}. BioBERT \cite{lee2020biobert} is a pre-trained language representation model specifically designed for biomedical text mining, while it enhanced performance in biomedical tasks such as named entity recognition (NER), relation extraction (RE), and question answering (QA) by adapting the general BERT model to biomedical corpora. Similarly, BioGPT \cite{luo2022biogpt}, a generative pre-trained transformer model, is tailored for biomedical text generation and mining. It outperformed previous models in tasks like relation extraction, QA, and document classification, highlighting its ability to handle domain-specific language and surpassing general models like GPT-2 in generating meaningful biomedical content. Although BioBERT and BioGPT have been pivotal in biomedical natural language processing, they are relatively small compared to the latest large-scale language models. BioBERT, based on the BERT base model, contains 110 million parameters, while BioGPT, built upon GPT-2, has 1.5 billion parameters. In contrast, advanced LLMs like the PaLM~\cite{chowdhery2023palm} and LLaMA~\cite{touvron2023llama} families feature tens to hundreds of billions of parameters, greatly exceeding earlier biomedical models in their capacity to handle complex medical tasks.

As a result, there is a growing trend to fine-tune larger LLMs for medical applications, given their potential to improve performance across various complex tasks. However, fine-tuning LLMs with hundreds of billions of parameters is computationally intensive. To address this challenge, Med-PaLM~\cite{singhal2023large} adapts the Flan-PaLM model~\cite{chung2024scaling} to the medical domain using instruction prompt tuning instead of full-model fine-tuning. This approach achieves state-of-the-art results in medical question answering, multiple-choice questions, and open-domain tasks~\cite{jin2021disease, pal2022medmcqa, jin2019pubmedqa}, leveraging models with 8B, 62B, and 540B parameters from the PaLM and Flan-PaLM families~\cite{chowdhery2023palm, chung2024scaling}. Additionally, PMC-LLaMA~\cite{wu2024pmc} is specifically designed for medical applications, incorporating biomedical and clinical data through a two-step training process: data-centric knowledge injection followed by medical-specific instruction tuning. This model is based on LLaMA-v2 with 13B parameters. Similarly, Me-LLaMA~\cite{xie2024me} introduces a family of medical LLMs using LLaMA-v2 models with 13B and 70B parameters, adopting a 4:1 ratio between medical and general domain data, and demonstrates significant performance improvements in both general and medical tasks.

\textbf{Efficient LLM Fine-tuning.}

Low-Rank Adaptation (LoRA) \cite{hu2021lora} has emerged as a highly efficient method for fine-tuning LLMs in resource-constrained environments, allowing for effective model specialization with minimal computational overhead. Rather than fine-tuning all parameters of a model, LoRA updates only a small subset by introducing low-rank matrices into each layer of the Transformer architecture. This approach dramatically reduces memory usage while maintaining high task-specific performance. By freezing the pre-trained model weights and adding trainable rank-decomposition matrices, LoRA significantly decreases the number of trainable parameters required for downstream tasks, making it particularly effective for scenarios with limited computational resources. 

Building on the strengths of LoRA, MOELoRA \cite{liu2024moe} is introduced as a parameter-efficient fine-tuning framework for multi-task medical applications. It leverages the mixture-of-experts (MOE) approach for multi-task learning and combines it with LoRA’s parameter efficiency. This hybrid method enables fine-tuning across diverse medical tasks while maintaining a low parameter footprint, ensuring scalability and adaptability in complex healthcare environments. Additionally, Le~\etal~\cite{le2024impact} have examined various LoRA adapter structures for clinical natural language processing, demonstrating their ability to rapidly and efficiently adapt LLMs for tasks such as narrative classification in clinical settings.

\textbf{Web-Based Medical Interfaces and Visualization.}

The integration of LLMs into web-based medical platforms is becoming increasingly common, driven by the need for user-friendly and accessible AI tools in healthcare. Applications such as ChatGPT~\footnote{https://chat.openai.com/}, Claude~\footnote{https://claude.ai/}, and Gemini~\footnote{https://gemini.google.com/} exemplify this trend by offering intuitive interfaces that enable both clinicians and patients to input medical data and receive diagnostic feedback in real time.

Web-based interfaces enhance accessibility and usability by simplifying interactions with LLMs. Users can enter symptoms and medical histories directly, receiving AI-generated diagnostic suggestions that support faster and more informed decision-making. For instance, ChatCAD~\cite{wang2024interactive} leverages the ChatGPT interface to integrate disease classification, lesion segmentation, and report generation into a single prompt-driven workflow. However, its inability to fine-tune the underlying model and dependence on an online service raise concerns about specialization and data privacy.

MedPaLM-2~\cite{singhal2023towards}, developed by Google, is a domain-specific LLM aimed at supporting clinical decision-making and medical information retrieval. Built on Google's Vertex AI platform, it employs advanced NLP techniques tailored for healthcare. Despite its potential, MedPaLM-2 is limited to selected institutions and lacks an open-access web interface, while also raising concerns about handling sensitive data securely in online environments.

In terms of visualization, RetainVis~\cite{kwon2018retainvis} employs bar, area, and circle charts to depict diagnosis distributions within electronic medical records. GLoRIA~\cite{huang2021gloria} focuses on visualizing attention maps in radiographic images, aiding interpretability of model outputs.


\section{Methodology}

In this section, we present an overview of the entire system, which is composed of two main components: the methodology used to develop our medical pre-diagnosis model, utilizing a LoRA-tuned LLaMA-v3 LLM, and a Flask-based web interface that supports both user input and chart-based output for differential diagnosis. Our approach aims to deliver high accuracy in pathology prediction and differential diagnosis while maintaining deployability within privacy-sensitive healthcare settings. The proposed method integrates LoRA fine-tuning, local deployment, explainability techniques, and a web-based user interface, offering an effective, secure, and transparent solution for supporting medical diagnosis.

\begin{figure}[t!]
    \centering
    \includegraphics[width = 0.9999\linewidth]{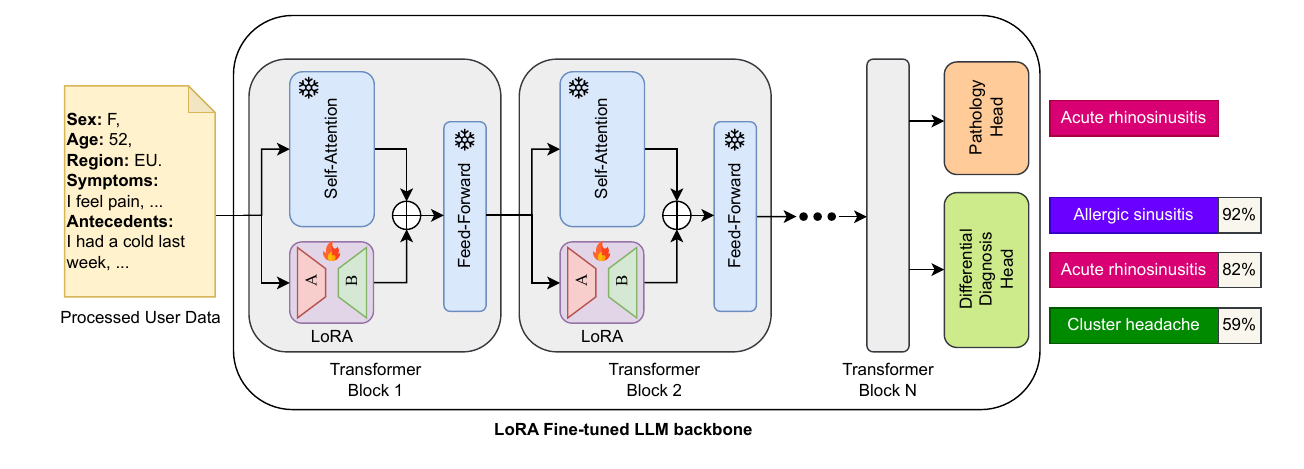}
    \caption{The proposed medical document analysis architecture for pathology and differential diagnosis.}
    \label{fig:arch}
\end{figure}

\subsection{Backbone LLM: LLaMA-v3}
We select the latest Meta's LLaMA-v3 model as the foundational LLM for our platform due to its strong performance, scalability, and adaptability, offering versatility for various applications. Compared to LLaMA-v2, which is used by PMC-LLaMA~\cite{wu2024pmc} and Me-LLaMA~\cite{xie2024me}, LLaMA-v3 demonstrates enhanced capability in understanding complex language patterns, which is essential for processing nuanced medical information and performing differential diagnosis. Specifically, we utilized the ``Meta-Llama-3.1-8B-Instruct'' variant in this paper, as illustrated using light blue rectangles in Fig.~\ref{fig:arch}.

\subsection{Fine-Tuning Strategy}


Aghajanyan~\etal~\cite{aghajanyan2020intrinsic} demonstrated that pre-trained language models have a low ``intrinsic dimension'', enabling efficient learning even when projected onto a smaller subspace. Building on this, LoRA~\cite{hu2021lora} hypothesizes that weight updates during adaptation also exhibit a low ``intrinsic rank''. We apply LoRA to fine-tune LLaMA-v3, as shown by the purple rectangles in Fig.~\ref{fig:arch}, alongside the frozen self-attention modules. This enables domain-specific adaptation for specialized tasks such as pathology prediction and differential diagnosis.


LLaMA-v3 is composed of an Embedding layer along with multiple self-attention and MLP blocks. To address the tasks of pathology prediction and differential diagnosis, we design a sequence classification pipeline by appending additional linear heads to the end of the LLaMA-v3 model. Our approach focuses on adapting LoRA specifically for the self-attention modules in each block, leaving the MLP modules unchanged. For a pre-trained LLaMA-v3 model, let $W_0 \in \mathbb{R}^{d \times k}$ represent the self-attention weight matrix. We constrain its update by expressing it as a low-rank decomposition, $W_0 + \Delta W = W_0 + BA$, where $B \in \mathbb{R}^{d \times r}$, $A \in \mathbb{R}^{r \times k}$, and the rank $r \ll \min(d, k)$. During training, $W_0$ remains fixed without gradient updates, while $A$ and $B$ contain the trainable parameters. Note that both $W_0$ and $\Delta W = BA$ are multiplied by the same input, and their resulting output vectors are added coordinate-wise. For $h = W_0 x$, our modified forward pass becomes:

\[
h = W_0 x + \Delta W x = W_0 x + B A x
\]
where $A$ is randomly Gaussian initialized and $B$ is zero initialized so as $\Delta W = BA$ is zero at the beginning of training.

\subsection{Web Interface}

\begin{figure}[ht!]
    \centering
    \includegraphics[width = 0.9999\linewidth]{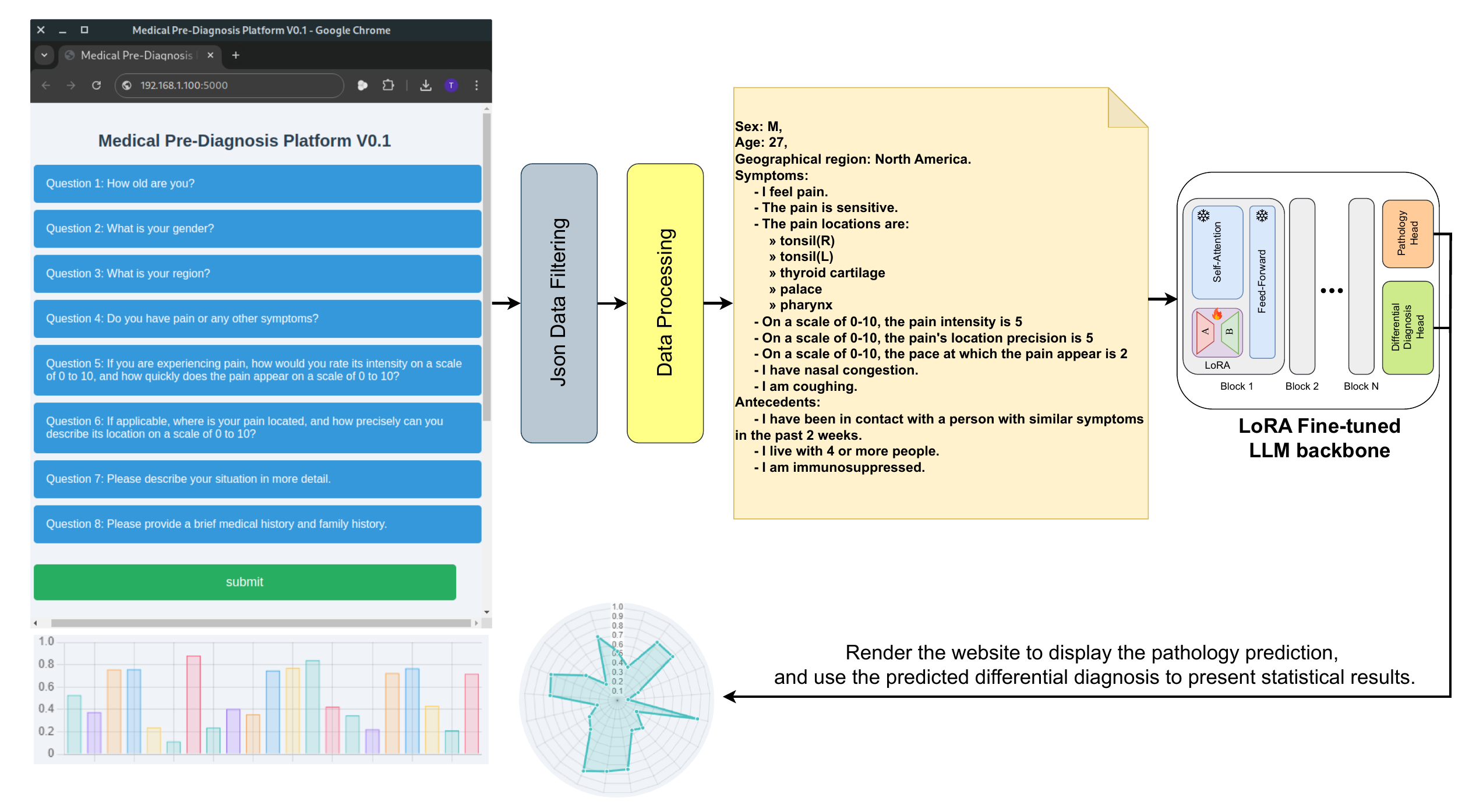}
    \caption{The proposed medical pre-diagnosis platform pipeline.}
    \label{fig:pipe}
\end{figure}

To ensure secure and private access, our LoRA fine-tuned LLM is deployed via a hospital-hosted web interface. Unlike general-purpose models like MedPaLM-2, our system is specialized for high-accuracy differential diagnosis and pathology pre-assessment. It provides patients with preliminary insights while awaiting formal diagnosis. The system pipeline is illustrated in Fig.~\ref{fig:pipe}. Our platform comprises a JavaScript-based front-end and a Flask back-end, enabling real-time data processing and visualization.

\textbf{Back-end with Flask:}

\begin{itemize}
    \item \textbf{Routing and APIs:} Flask manages user interactions through API endpoints. User responses to medical questions are sent as POST requests and formatted for model input.
    \item \textbf{Data Processing:} Input data is validated and transformed using regular expressions before being passed to the neural network.
    \item \textbf{Model Inference:} The fine-tuned LLM runs on a GPU-enabled server, and Flask orchestrates model inference and session management.
\end{itemize}

\textbf{Front-end with JavaScript and Chart.js:}

\begin{itemize}
    \item \textbf{User Interaction:} JavaScript handles form inputs, validates data, and communicates with the back-end asynchronously.
    \item \textbf{Visualization:} Chart.js renders real-time charts (bar and radar) based on the model's outputs, enhancing interpretability.
\end{itemize}

\textbf{Front-End and Back-End Communication Workflow:}
\begin{itemize}
\item User Interaction: The user inputs data (e.g., age, sex, symptoms, medical history, etc.) into the form on the web page.
\item Data Submission: JavaScript sends the form data to the Flask back-end.
\item LLM Processing: The Flask server forwards the user data to our LLM, which analyzes it and makes predictions.
\item Result Generation: The Flask back-end prepares the results and sends them back to the front-end.
\item Visualization: JavaScript processes the returned results and uses Chart.js to create a dynamic chart, presenting the results to the user in a clear and interactive manner.
\end{itemize}

\subsection{Qualitative Result}

\begin{figure}[ht!]
    \centering
    \begin{subfigure}{0.49\textwidth}
        \includegraphics[width=0.8\linewidth]{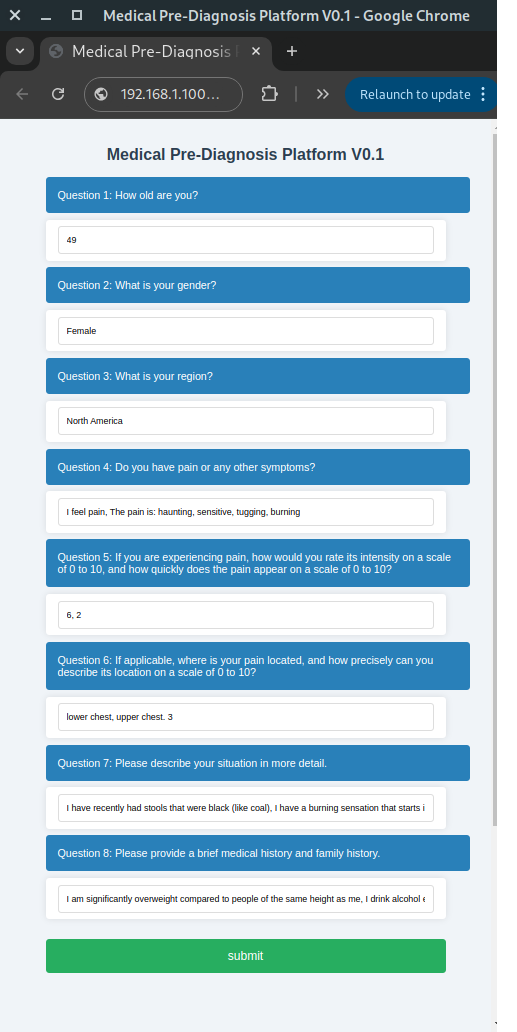}
        \caption{User input}
        \label{fig:sub1}
    \end{subfigure}
    \vspace{0.8em}
    \begin{subfigure}{0.49\textwidth}
        \includegraphics[width=0.8\linewidth]{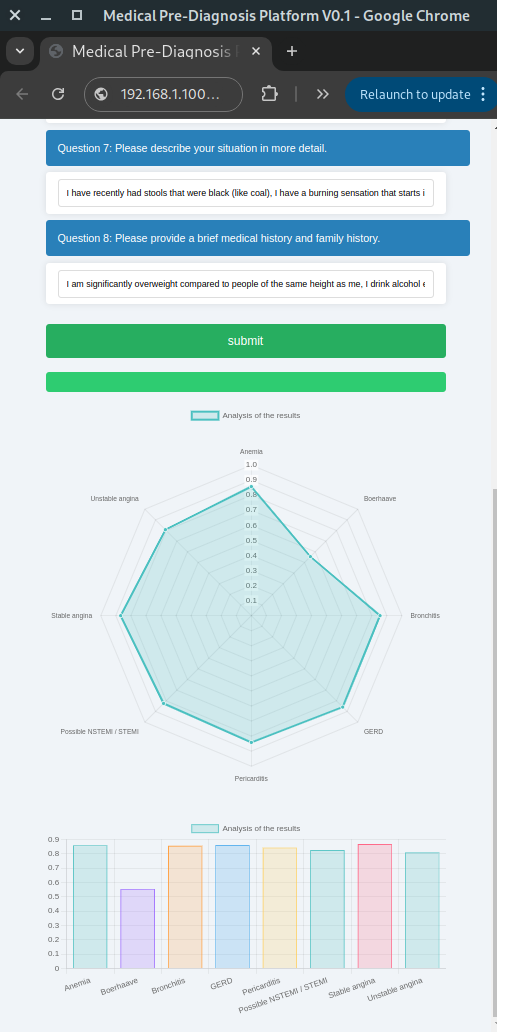}
        \caption{Output of differential diagnosis results}
        \label{fig:sub2}
    \end{subfigure}

    \caption{Visualization of the website platform.}
    \label{fig:web}
\end{figure}

The website user interface is displayed in Fig.~\ref{fig:web}. Users are presented with 8 basic medical-related questions, as listed in Tab.~\ref{tab:ques}. Upon clicking the submit button, a progress bar appears, indicating the inference process by an internal LLM. The results are then displayed in radar and bar charts, as shown in the bottom-right corner of Fig.~\ref{fig:web}.

\vspace{-0.5cm}
\begin{table}[ht!]
    \caption{Medical question form on the web interface.}
    
    \label{tab:ques}
    \centering
    \small
    \scalebox{0.8}{
    \begin{tabular}{c|l}
    \toprule
    \textbf{ID} & \textbf{Question Content}\\
    \specialrule{0.5pt}{1pt}{1pt}
    1 & How old are you?\\
    \specialrule{0.5pt}{1pt}{1pt}
    2 & What is your gender?\\
    \specialrule{0.5pt}{1pt}{1pt}
    3 & What is your region?\\
    \specialrule{0.5pt}{1pt}{1pt}
    4 & Do you have pain or any other symptoms?\\
    \specialrule{0.5pt}{1pt}{1pt}
    \multirow{2}{*}{5} & If you are experiencing pain, how would you rate its intensity on a scale\\
    & of 0 to 10, and how quickly does the pain appear on a scale of 0 to 10?\\
    \specialrule{0.5pt}{1pt}{1pt}
    \multirow{2}{*}{6} &  If applicable, where is your pain located, and how precisely can you\\
    & describe its location on a scale of 0 to 10?\\
    \specialrule{0.5pt}{1pt}{1pt}
    7 & Please describe your situation in more detail.\\
    \specialrule{0.5pt}{1pt}{1pt}
    8 & Please provide a brief medical history and family history.\\
    \bottomrule
    \end{tabular}
    }
\end{table}

\vspace{-0.5cm}



\section{Quantitative Performance Evaluation}

\subsection{DDXPlus Dataset}
\label{sec:data}

The DDXPlus dataset is one of the largest publicly available datasets designed for differential diagnosis. It features around 1.3 million synthetic patient records, each including information on symptoms, medical history, and ground truth conditions, which are vital for training models in automated differential diagnosis. The dataset covers 49 pathologies across a range of age groups, genders, and patients with varied medical backgrounds. Since the patient's true pathology is not always ranked first in the differential diagnosis, we do not use the probability of each pathology for training. Instead, we rely on the variable-length list of ground truth differential diagnoses. 

Thus, we concentrate on two key tasks: pathology prediction, which involves identifying the single correct pathology, and differential diagnosis, which entails predicting a variable-length list of potential pathologies.

\subsection{Implementation Details}

We train the model separately for the pathology prediction and differential diagnosis tasks, with each task requiring 108 hours to complete a single epoch using a batch size of 2. The LoRA parameters include an alpha of 16, a dropout rate of 0.1, and a rank $r$ set to 4. The pathology prediction task is trained for one epoch, while the differential diagnosis task is trained for two epochs. Evaluating the model on the test set takes 7.5 hours to generate the results. All experiments are conducted on a single NVIDIA A40 GPU.

\subsection{Metrics}
We evaluate the effectiveness of our method using several widely recognized performance metrics: Accuracy, Precision, Recall, F1 score, and Ground Truth Pathology Accuracy (GTPA)~\cite{fansi2022ddxplus}. Accuracy is defined as:

\[
Accuracy = \frac{TP + TN}{TP + FP + TN + FN}
\]
where $TP$ (True Positives), $FP$ (False Positives), $TN$ (True Negatives), and $FN$ (False Negatives) are the counts of each classification outcome. This metric represents the overall proportion of correct predictions.

Precision is defined as:
\[
Precision = \frac{TP}{TP + FP}
\]

Recall is defined as:
\[
Recall = \frac{TP}{TP + FN}
\]

The F1 score is calculated as:
\[
F1\_score = 2*\frac{Precision*Recall}{Precision + Recall}
\]

The GTPA metric assesses whether the agent's predicted differential diagnosis includes the true pathology $p^i$ of each patient:
\[
GTPA = \frac{1}{|D|} \sum_{i=1}^{|D|} \mathds{1} [p^i \in \hat{Y}_i]
\]
where $|D|$ is the total number of patients, $\hat{Y}_i$ is the set of predicted pathologies for the $i$-th patient, and $\mathds{1}$ is the indicator function that equals 1 if its argument is true and 0 otherwise.

\subsection{Performance on Pathology Prediction}

The pathology prediction task involves predicting only the single most likely disease. From Tab.~\ref{tab:patho}, it is evident that our proposed method achieves approximately $95\%$ for precision, recall, and F1 score, along with an accuracy of $99.81\%$. Although this accuracy is slightly lower than that of the DDxT method, which achieved $99.98\%$, our model's performance remains suitable for practical applications due to its high degree of accuracy. It is important to note that the DDxT method did not report precision, recall, or F1 scores for the pathology prediction task.

\vspace{-0.5cm}
\begin{table}[ht!]
    \caption{Comparison with the State of the Art for Pathology Prediction using the DDXPlus Dataset.}
    \label{tab:patho}
    \small
    \begin{minipage}{\textwidth}
     \centering
    \begin{tabular}{ccccc}
    \toprule
    \textbf{Method} & \textbf{Accuracy} & \textbf{Precision} & \textbf{Recall} & \textbf{F1}\\
    \midrule
    BASD~\cite{luo2021knowledge} & 97.15 & $-$ & $-$ & $-$\\
    AARLC~\cite{yuan2024efficient} & 99.21 & $-$ & $-$ & $-$\\
    DDxT~\cite{alam2023ddxt} & \textbf{99.98} & $-$ & $-$ & $-$\\
    \midrule
    \textbf{Proposed} & 99.81 & \textbf{96.54} & \textbf{94.34} & \textbf{94.81}\\
    \bottomrule
    \end{tabular}
    \end{minipage}
    
\end{table}
\vspace{-0.5cm}

\subsection{Performance on Differential Diagnosis}

Differential diagnosis involves predicting a variable-length set of diseases with corresponding probabilities. Tab.~\ref{tab:diag} compares our proposed method with three baseline methods for the differential diagnosis prediction task.

Our method significantly outperforms the others. It achieves the highest GTPA of $99.94\%$, surpassing BASD at $99.30\%$ and AARLC at $99.92\%$, indicating a more consistent agreement with the ground truth pathology. Our method also attains a notable accuracy of $99.46\%$; accuracy values for BASD, AARLC, and DDxT were not reported.

In terms of precision, our model achieves $98.18\%$, outperforming BASD at $88.34\%$, AARLC at $69.53\%$, and DDxT at $94.84\%$. This suggests our method better avoids false positive diagnoses, providing more reliable results.

Our method attains a recall of $97.91\%$, comparable to AARLC at $97.73\%$ and slightly higher than DDxT at $94.65\%$, demonstrating competitive sensitivity. BASD, by contrast, has a lower recall of $85.03\%$, indicating it misses a greater proportion of true positive cases.

Finally, our model achieves the highest F1 score of $98.01\%$, indicating an optimal balance between precision and recall and outperforming the state of the art by a large margin.

\begin{table}[ht!]
    \caption{Comparison with the State of the Art for Differential Diagnosis using the DDXPlus Dataset.}
    \label{tab:diag}
    \centering
    \small
    \scalebox{0.99}{
    \begin{tabular}{cccccc}
    \toprule
    \textbf{Method} & \textbf{GTPA} & \textbf{Accuracy} & \textbf{Precision} & \textbf{Recall} & \textbf{F1}\\
    \midrule
    BASD~\cite{luo2021knowledge} & 99.30 & $-$ & 88.34 & 85.03 & 83.69\\
    AARLC~\cite{yuan2024efficient} & 99.92 & $-$ & 69.53 & 97.73 & 78.24\\
    DDxT~\cite{alam2023ddxt} & $-$ & $-$ & 94.84 & 94.65 & 94.72\\
    \midrule
    \textbf{Proposed} & \textbf{99.94} & \textbf{99.46} & \textbf{98.18} & \textbf{97.91} & \textbf{98.01}\\
    \bottomrule
    \end{tabular}
    }
\end{table}

\subsection{Explainability Insights}

\begin{figure}[!t]
    \centering
    \begin{minipage}[t]{0.36\textwidth}
        \centering
        \begin{subfigure}{\linewidth}
            \includegraphics[width=\linewidth]{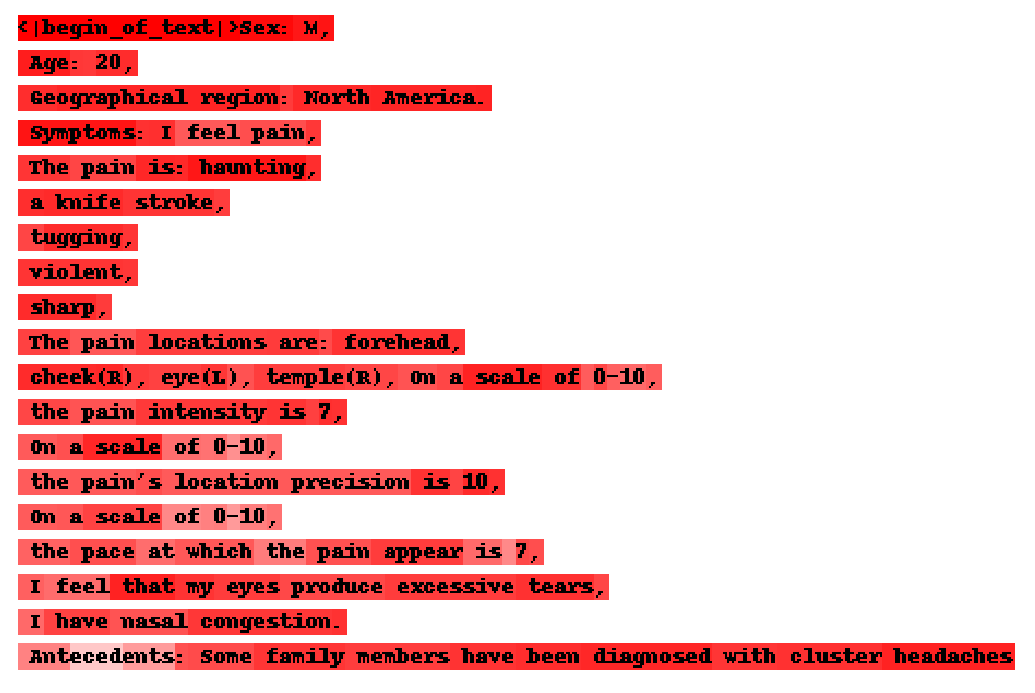}
            \caption{Shallow Layer}
            \label{fig:sub1}
        \end{subfigure}
        \vspace{0.5em}
        \begin{subfigure}{\linewidth}
            \includegraphics[width=\linewidth]{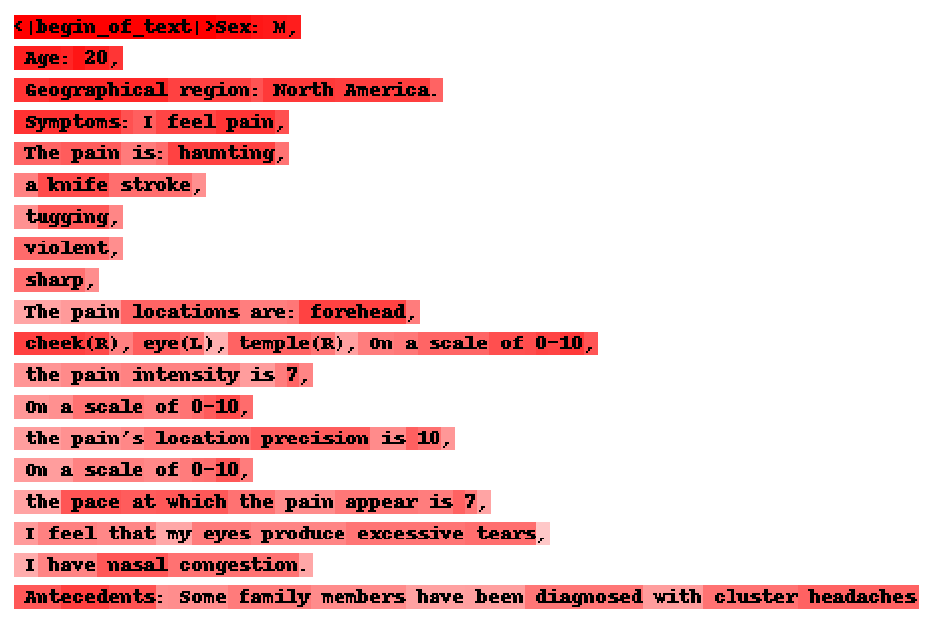}
            \caption{Middle Layer}
            \label{fig:sub2}
        \end{subfigure}
        \vspace{0.2em}
        \begin{subfigure}{\linewidth}
            \includegraphics[width=\linewidth]{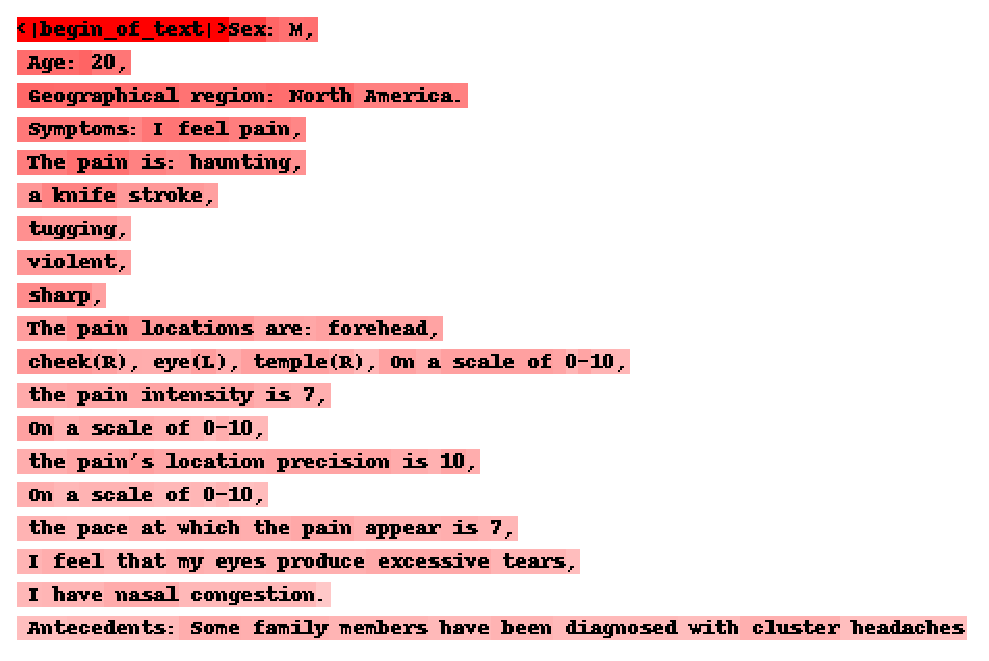}
            \caption{Deep Layer}
            \label{fig:sub3}
        \end{subfigure}
        \caption{Self-Attention visualization for Shallow, Middle and Deep layers in LoRA-tuned LLaMA-v3 model for correct pathology prediction for ``Cluster headache''.}
        \label{fig:vis}
    \end{minipage}
    \hfill
    \vspace{0.2cm}
    \begin{minipage}[t]{0.62\textwidth}
        \centering
        \begin{subfigure}{\linewidth}
            \includegraphics[width=\linewidth]{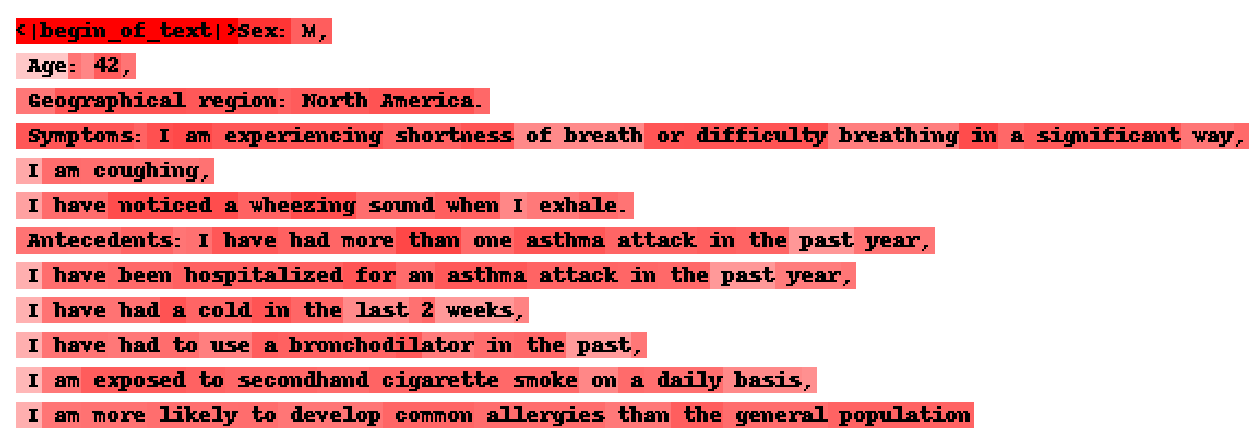}
            \caption{Shallow Layer}
            \label{fig:sub1_error}
        \end{subfigure}
        \vspace{0.5em}
        \begin{subfigure}{\linewidth}
            \includegraphics[width=\linewidth]{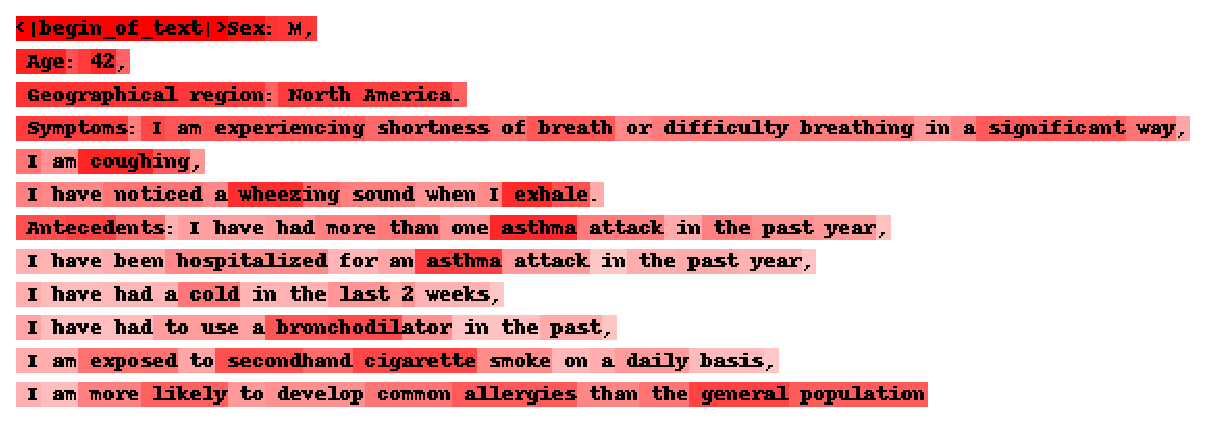}
            \caption{Middle Layer}
            \label{fig:sub2_error}
        \end{subfigure}
        \vspace{0.2em}
        \begin{subfigure}{\linewidth}
            \includegraphics[width=\linewidth]{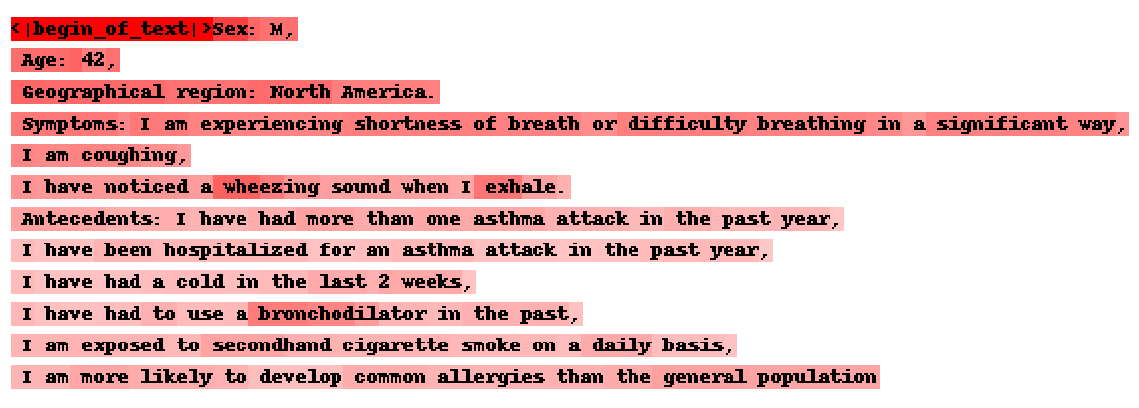}
            \caption{Deep Layer}
            \label{fig:sub3_error}
        \end{subfigure}
        \caption{A failure case for self-Attention visualization in LoRA-tuned LLaMA-v3 model. The groundtruth pathology is ``Bronchospasm / acute asthma exacerbation'', but the prediction is ``Acute COPD exacerbation / infection''.}
        \label{fig:vis_error}
    \end{minipage}
\end{figure}

In Fig.~\ref{fig:vis}, we illustrate the consistent self-attention patterns across the shallow, middle, and deep layers of our LoRA-tuned LLaMA-v3 model. In the shallow layer, depicted in (a) of Fig.~\ref{fig:vis}, the self-attention module attends broadly to all input tokens, supporting comprehensive feature extraction. In the middle layer, shown in (b), the self-attention selectively focuses on specific elements, such as the keywords ``forehead'', ``cheek(R)'', ``eye(L)'', and ``temple(R)'', indicating pain locations, along with the pain intensity value ``7''. Finally, in the deep layer, visualized in (c), self-attention concentrates predominantly on the initial tokens, aligning with the sequence classification objective, where the information is distilled into the special token ``<|begin\_of\_text|>''.

In Fig.~\ref{fig:vis_error}, we present a failure case illustrating the self-attention patterns in the shallow, middle, and deep layers of our LoRA-tuned LLaMA-v3 model. In the shallow layer, depicted in (a) of Fig.~\ref{fig:vis_error}, the behavior resembles that of the correct case in Fig.~\ref{fig:vis}: the model attends broadly to all input tokens for feature extraction but interestingly it excludes redundant words like ``Age'' and ``I''. In the middle layer, shown in (b), the self-attention focuses on key symptoms such as ``wheezing'', ``exhale'', and ``cold''. Notably, despite a strong focus on the term ``asthma'', the model still incorrectly predicts ``Acute COPD exacerbation / infection'' instead of the ground truth ``Bronchospasm / acute asthma exacerbation''. In the final layer, visualized in (c), the attention attempts to concentrate on the initial tokens, aligning with the sequence classification task. However, key terms like ``wheezing'' and ``bronchodilator'' still retain attention, indicating inconsistency in feature consolidation. These subtle inconsistencies in the attention patterns ultimately lead to unstable pathology predictions, resulting in failure.

\begin{figure}[!t]
    \centering
    \begin{subfigure}{0.35\textwidth}
        \includegraphics[width=\linewidth]{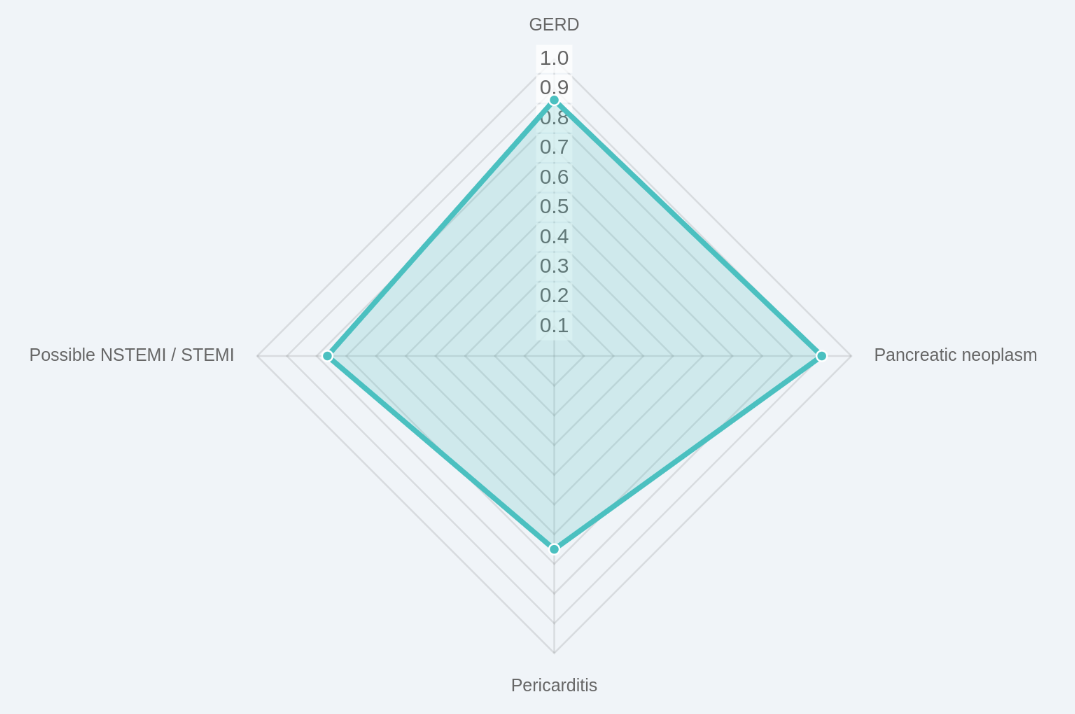}
        \caption{Radar Chart}
        \label{fig:sub1}
    \end{subfigure}
    \hfill
    \begin{subfigure}{0.6\textwidth}
        \includegraphics[width=\linewidth]{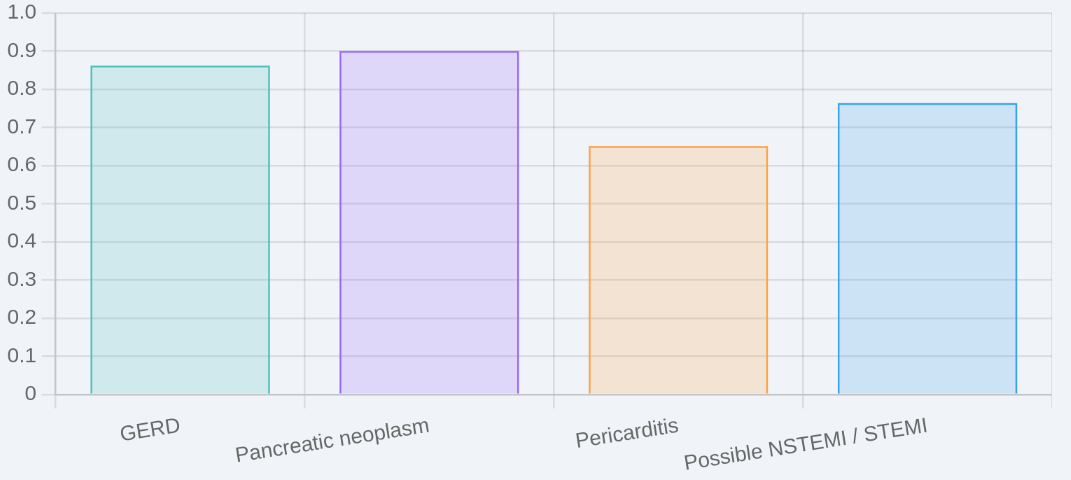}
        \caption{Bar Chart}
        \label{fig:sub2}
    \end{subfigure}

    \caption{A failure case where the differential diagnosis predicts only 4 cases, while the ground truth includes an additional 3 cases.}
    \label{fig:chart_fail}
\end{figure}

\begin{figure}[!t]
    \centering
    \begin{subfigure}{0.35\textwidth}
        \includegraphics[width=\linewidth]{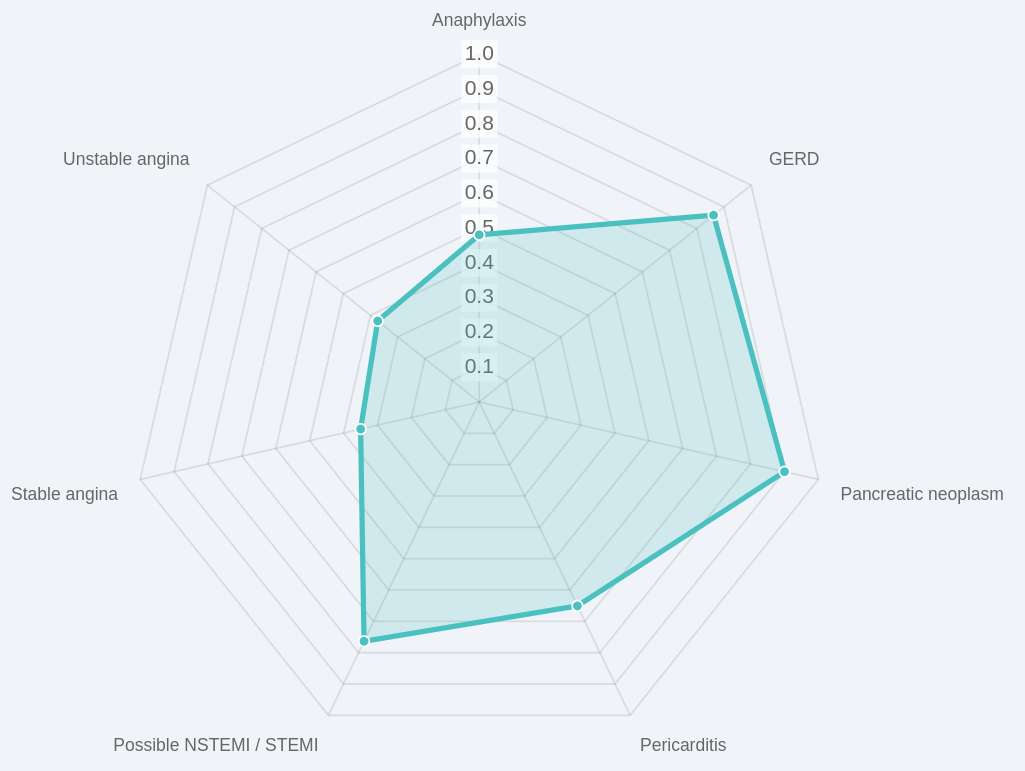}
        \caption{Radar Chart}
        \label{fig:sub1}
    \end{subfigure}
    \hfill
    \begin{subfigure}{0.6\textwidth}
        \includegraphics[width=\linewidth]{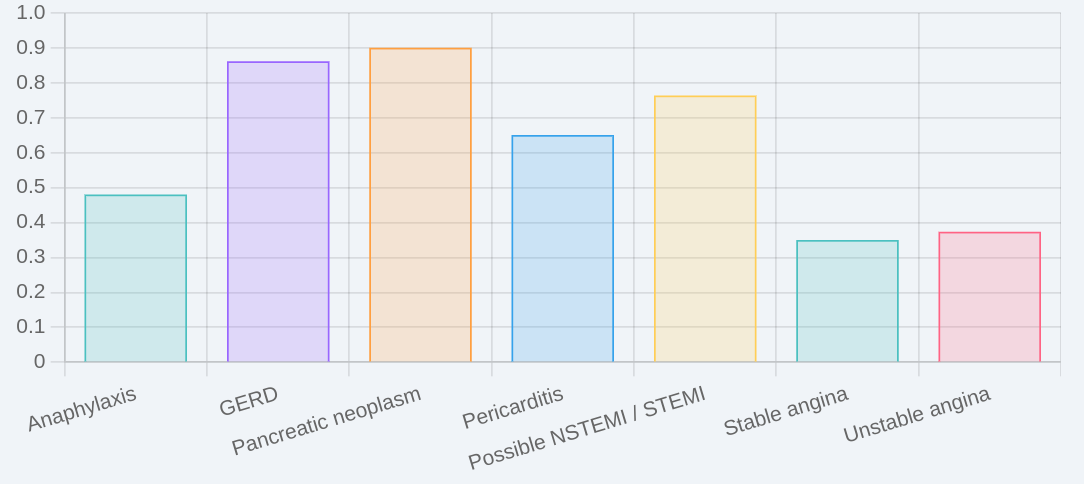}
        \caption{Bar Chart}
        \label{fig:sub2}
    \end{subfigure}

    \caption{By lowering the prediction threshold from 0.5, three additional cases are included, resulting in the total differential diagnosis matching the ground truth exactly.}
    \label{fig:chart_fail_corr}
\end{figure}

\begin{figure}[!t]
    \centering
    \begin{subfigure}{0.45\textwidth}
        \includegraphics[width=\linewidth]{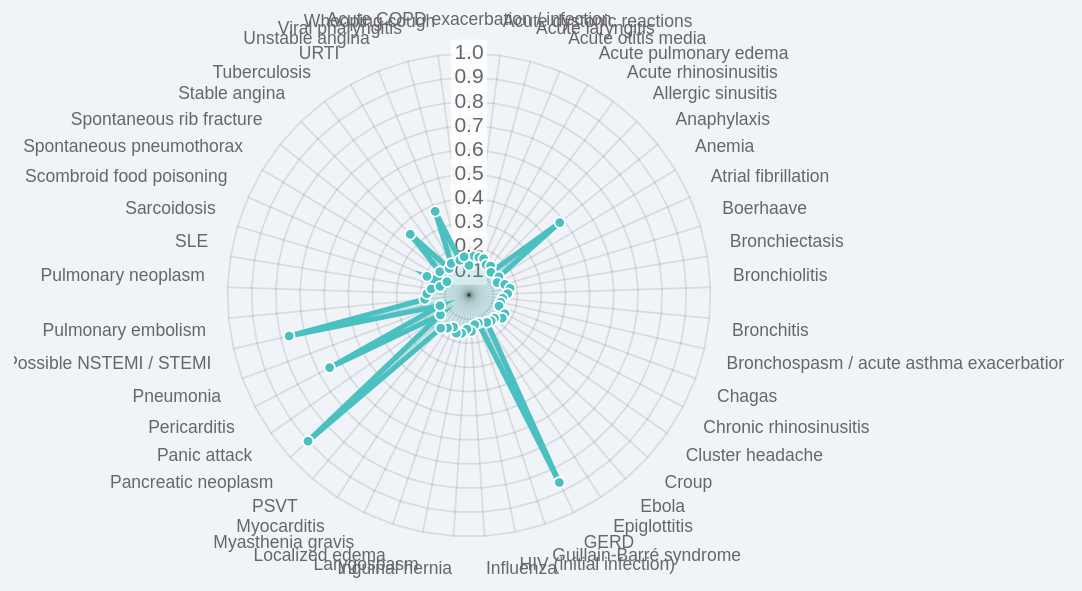}
        \caption{Radar Chart}
        \label{fig:sub1}
    \end{subfigure}
    \hfill
    \begin{subfigure}{0.5\textwidth}
        \includegraphics[width=\linewidth]{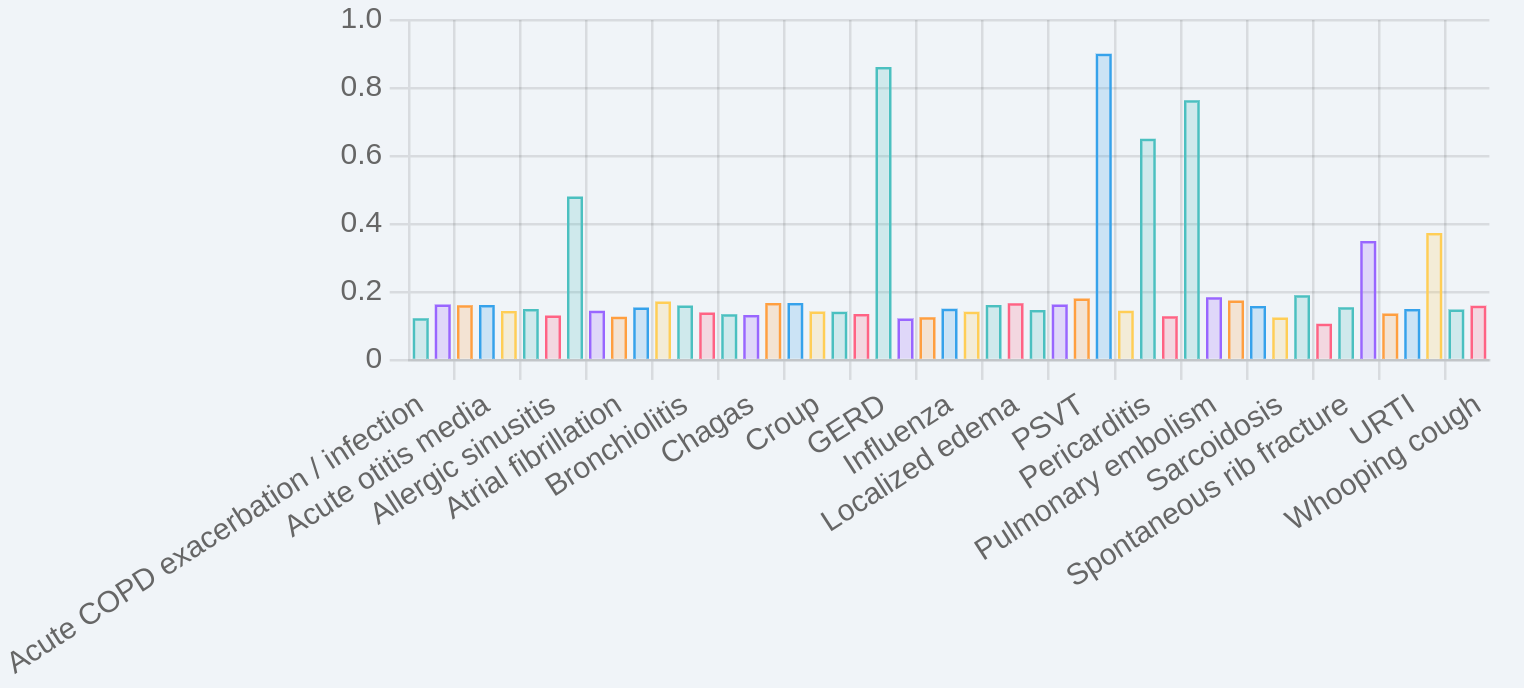}
        \caption{Bar Chart}
        \label{fig:sub2}
    \end{subfigure}

    \caption{Displaying all 49 disease categories for differential diagnosis. }
    \label{fig:chart_all}
\end{figure}

\subsection{Discussion of the Failure Prediction Charts}

Although our proposed model has achieved over 99\% accuracy, it is essential to analyze failure cases due to the critical nature of medical applications. Fig.~\ref{fig:chart_fail} presents the prediction results. In this analysis, we apply a sigmoid function to the differential diagnosis head and use a threshold of 0.5 to identify high-probability pathology categories as predictions. This approach results in 4 predicted cases, whereas the ground truth contains a total of 7 cases, leaving 3 cases unaccounted for.

By simply lowering the prediction threshold from 0.5 to 0.35 as illustrated in Fig.~\ref{fig:chart_fail_corr}, we can include 3 additional cases, perfectly matching the ground truth differential diagnosis. This indicates that our proposed method can still address these failure cases, with the discrepancy being due to the threshold setting. As mentioned in Section~\ref{sec:data}, the patient's true pathology is not always ranked first in the differential diagnosis in the DDXPlus dataset, so we do not use the probability of each pathology for training. As a result, we lack probability-based ground truth for training the LLM. It is noteworthy that even without supervised probability, the model can inherently learn probability estimates based on patient symptoms and basic information.

Furthermore, we reduce the threshold to 0 to include all pathology categories, resulting in predictions for all 49 categories, as shown in Fig.~\ref{fig:chart_all}. This visualization reveals that the top 7 categories, which align with the ground truth differential diagnosis, include 4 categories with probabilities above 0.5 and 3 categories with probabilities between 0.35 and 0.5. These 7 categories exhibit relatively higher probabilities compared to the remaining 42 categories, all of which have probabilities below 0.2.


\section{Conclusion}

In this paper, we introduce an LLM-driven method for analyzing medical documents that delivers reliable pathology predictions and differential diagnoses. Additionally, we present a user-friendly, web-based pre-diagnosis platform. Our experiments show that the proposed approach performs on par with or even exceeds state-of-the-art methods. Moreover, our explainability experiments—where we visualize self-attention maps from shallow, middle, and deep layers—reveal consistent logical patterns. These insights help us distinguish between correct and incorrect predictions, thereby enhancing the model’s interpretability and reliability.

By leveraging minimalist design principles for both front-end and back-end development, our platform offers an intuitive interface that delivers clear and actionable visualizations for medical document analysis. This design not only streamlines the extraction and interpretation of critical clinical data but also positions the system as a vital asset for hospitals and healthcare providers. Looking ahead, we aim to further enhance its usability and functionality, enabling rapid analysis of patient records to facilitate early detection and timely intervention. Ultimately, our work aspires to set a new standard in medical document analysis and pre-diagnosis, contributing to improved patient outcomes and more efficient clinical workflows.

\section{Acknowledgments}
This work has been partially supported by the Beatriu de Pinós del Departament de Recerca i Universitats de la Generalitat de Catalunya (2022 BP 00256), the predoctoral program AGAUR-FI ajuts (2025 FI-2 00470) Joan Oró, which is backed by the Secretariat of Universities and Research of the Department of Research and Universities of the Generalitat of Catalonia and the European Social Plus Fund, the Spanish projects PID2021-126808OB-I00 (GRAIL) and CNS2022-135947 (DOLORES), grant PID2021-128178OB-I00 funded by MCIN/AEI/10.13039/501100011033, ERDF ``A way of making Europe'', the project PID2023-146426NB-100 funded by MCIU/AEI/10.13039/501100011033 and FSE+, and European Lighthouse on Safe and Secure AI (ELSA) from the European Union’s Horizon Europe programme under grant agreement No 101070617.

\bibliographystyle{splncs04}
\bibliography{bib}

\end{document}